\documentclass[journal]{IEEEtran}

\usepackage{amsmath}
\usepackage{amssymb}
\usepackage{algorithm}
\usepackage{caption}
\usepackage{graphicx}
\usepackage{times}

\newcommand\etal{{et~al.}}
\newcommand\ie{{i.e.}}

\newcommand\eg{{e.g.}}

\ifCLASSINFOpdf
\else
\fi

\begin{document}

\title{Progressive Scene Text Erasing with Self-Supervision}

\author{Xiangcheng Du, Zhao Zhou, Yingbin Zheng, Xingjiao Wu, Tianlong Ma, Cheng Jin
\thanks{Xiangcheng Du, Zhao Zhou, Xingjiao Wu, Tianlong Ma, and Cheng Jin are with School of Computer Science, Fudan University, Shanghai, China. Yingbin Zheng is with Videt Technology, Shanghai, China. Xiangcheng Du and Zhao Zhou contributed equally to this work.} 
\thanks{Corresponding author: Cheng Jin.}
}

\maketitle

\begin{abstract}
Scene text erasing seeks to erase text contents from scene images and current state-of-the-art text erasing models are trained on large-scale synthetic data. Although data synthetic engines can provide vast amounts of annotated training samples, there are differences between synthetic and real-world data. In this paper, we employ self-supervision for feature representation on unlabeled real-world scene text images. A novel pretext task is designed to keep consistent among text stroke masks of image variants. We design the Progressive Erasing Network in order to remove residual texts. The scene text is erased progressively by leveraging the intermediate generated results which provide the foundation for subsequent higher quality results. Experiments show that our method significantly improves the generalization of the text erasing task and achieves state-of-the-art performance on public benchmarks.
\end{abstract}

\begin{IEEEkeywords}
Scene text erasing, self-supervision, Progressive Erasing Network, text stroke mask.
\end{IEEEkeywords}

\IEEEpeerreviewmaketitle

\section{Introduction}
\label{sec:intro}
\begin{figure}[t]
    \centering
    \includegraphics[width=.99\linewidth]{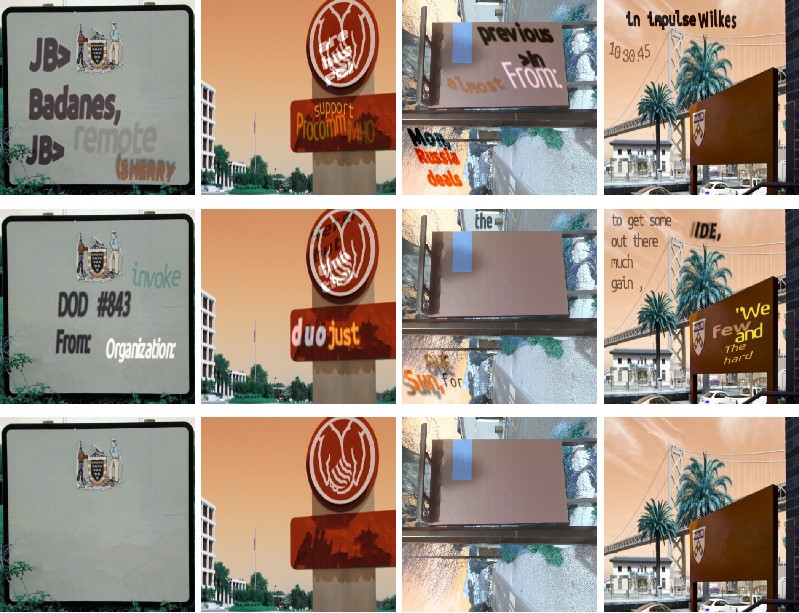}\\
    \vspace{-0.03in}
    {\footnotesize{(a) Synthetic text images}}\\
    \vspace{0.05in}
    \includegraphics[width=.99\linewidth]{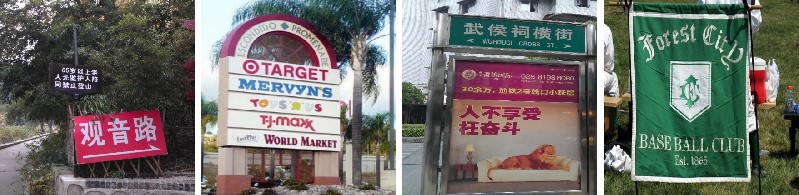}\\
    \vspace{-0.03in}
    {\footnotesize{(b) Real-world scene text images}}
    \caption{The synthetic text images (row 1, row 2) are generated by pasting random texts into normal scene images (row 3), which are usually unnatural compared with the real-world scene text images.}
    \vspace{-0.2in}
    \label{fig:synthetic_real-world}
\end{figure}

Scene text erasing has drawn remarkable attention in computer vision due to its importance in various real-world applications such as text transfer~\cite{azadi2018multi}, image editing~\cite{wu2019editing}, and image reconstruction~\cite{kao2019patch}. The main purpose of the task is to conceal text information that appears in scene images. The essence of achieving this goal is to distinguish text from background information, which needs to operate at the pixel level of the images. Therefore, a tremendous amount of training samples that contain text instances and corresponding background images without text are required to obtain a high-performance erasing model. Although the public real-world dataset such as SCTU-EnsText~\cite{liu2020erasenet} is available for model learning, the annotations are manually erased with Photoshop which is time-consuming for the complex scene images. Therefore, many previous methods (\eg, \cite{tursun2019mtrnet,tursun2020mtrnet++,tang2021stroke}) attempted to use data synthetic technology~\cite{gupta2016synthetic} to obtain numerous annotated data and further the trained models. However, the distribution inconsistency between synthetic data and real-world images still exists. As shown in Fig~\ref{fig:synthetic_real-world}, the font of synthetic samples is based on the standard font types, and the text rendered in the background is sometimes unnatural and unrealistic, while the real-world examples are with richer contexts. The bias between training samples will lead to the poor robustness of the text-erasing model.

To mitigate this issue, our intuition is to learn more sensitive feature representation on real-world training samples in a self-supervised learning manner. Self-supervised learning has demonstrated significant improvements in several computer vision applications, including image classification, object detection, and segmentation. The key of self-supervision is treating original data as supervised information to obtain learned representation and apply it for downstream tasks without manually annotating ground-truth. To this end, we introduce a novel pretext task that can utilize supervised information in large-scale unlabeled real-world scene images based on text stroke masks. Our method generates different variants for each original image such as contrast, color, and sharpness, the image variants preserve the original textual information. For the text erasing task, the generated final results are to fill the text content with background information. Therefore, text stroke masks created by subtracting results from corresponding augmented image variants need to be as consistent as possible as shown in Fig.~\ref{fig:pretext}(b). Through this, the text erasing model is guided to learn the robust feature representations in various unlabeled real data, which improves the network generalization.

Although self-supervision boosts the performance compared with the methods trained in synthetic datasets, some texts will remain during the single text erasing process. Consequently, we introduce a progressive scene text erasing framework to decompose the erasing process into multiple stages. The results of each stage are used to guide the execution of the next stage. We first perform a rough text erasing and then adopt continuous erasing to progressively achieve accurate results. In each iteration, the current image is generated based on the previously text-removed image to get one step closer to the optimal one. In this way, the difficulty of each step is intrinsically mitigated and residual text can be eliminated in a progressive manner. Therefore, a series of erasing processes can effectively improve the robustness of the model to large variations of scene text images.

To summarize, the major contributions of this paper are:
\begin{itemize}
    \item We provide a novel framework with self-supervision for text erasing, in which a large number of unlabeled scene images are used for robust feature representation learning.
    \item We design a new pretext task for the text erasing based on text stroke masks calculated from image variants which optimizes the network with implicit knowledge.
    \item The proposed Progressive Erasing Network considers the text erasing with learned strokes in a progressive manner, which clearly boosts the scene text erasing results.
    \item We prove the superiority of the proposed method on public benchmarks compared with the existing methods.
\end{itemize}

The rest of this paper is organized as follows. Section~\ref{sec:related} introduces the background of scene text erasing and the related tasks. Section~\ref{sec:Framework} discusses the model design and network architecture in detail. Section~\ref{sec:training} describes the training details of Progressive Erasing Network. In Section~\ref{sec:exp}, we demonstrate the qualitative and quantitative study of the framework. Finally, we conclude our work in Section~\ref{sec:conclusion}.

\section{Related Work}
\label{sec:related}

\subsection{Scene Text Erasing}

Existing approaches of scene text erasing can be classified into two major categories: one-stage and two-stage methods. The one-stage methods treat scene text erasing problem as image transformation inspired by generative adversarial network~\cite{goodfellow2014generative}. Nakamura~\etal~\cite{nakamura2017scene} was the first to successfully design a scene text eraser to ensconce the text in the wild which divides the image into small patches and erases the text respectively. After that, EnsNet~\cite{zhang2019ensnet} employed an end-to-end network that uses a cGAN~\cite{mirza2014conditional} with multiple loss functions to ensconce the text. MTRNet~\cite{tursun2019mtrnet} introduced auxiliary masks to stabilize the performance of the eraser. EraseNet was proposed in~\cite{liu2020erasenet} which employs a coarse-to-fine strategy with an additional segmentation head to boost the superiority of the model. The two-stage methods decompose the text erasing task into two subtasks: text detection and background inpainting. The text detection module predicts the text regions that need to be erased. The background inpainting module repaints those regions using an inpainting network. Zdenek~\etal~\cite{zdenek2020erasing} trains scene text detector and inpainting network separately. Training a scene text detector requires less expensive annotation than training a scene text eraser using the corresponding image pairs because only bounding boxes of text regions are required. The inpainting network is trained on scene images with random masks and therefore it is expected to be suitable for filling backgrounds. Tang~\etal~\cite{tang2021stroke} proposed a practical text erasing method using cropped text images instead of the entire image. Besides, the network contains a stroke prediction module that generates pixel-level masks of text.
\subsection{Self-Supervised Learning}
Recently, many self-supervised learning methods for visual feature learning have been developed without using any annotated labels~\cite{korbar2018cooperative,fernando2017self,buchler2018improving,huang2020learning}. Self-supervised learning is generally conducted in two stages: 1) Pretrain the model with a pretext task. 2) Using pretrained weights for initialization, train the model for the main task. These learned representations are intended to be useful for a wide range of downstream tasks. Numerous such pretext tasks have been proposed including predicting the relative location of patches in images~\cite{doersch2015unsupervised}, learning to match tracked patches~\cite{wang2015unsupervised}, predicting the colors in a grayscale image~\cite{zhang2016colorful} and filling in missing parts of images~\cite{pathak2016context}. RotNet~\cite{gidaris2018unsupervised} is a widely-used method, which predicts the rotation of images as a pretext task. The task is simple: rotate input images at 0, 90, 180, and 270 degrees, and the model recognizes the rotation applied to the image. These tasks are manually designed by experts to ensure that the learned representations are useful for downstream tasks like object detection, image classification, semantic segmentation, and instance discrimination. Dating back to~\cite{dosovitskiy2014discriminative}, the task of instance discrimination involves treating an image and it is transformed versions as one single class.
\subsection{Progressive Network}
Progressive processing is a popular strategy that is frequently employed in various tasks, such as scene text detection~\cite{wang2019shape}, scene text recognition~\cite{zhan2019esir,gao2020progressive}, RGB-D salient object detection~\cite{chen2020progressively}, and image super-resolution~\cite{saharia2021image}. PSENet~\cite{wang2019shape} generates the different scale of kernels for each text instance and gradually expands the minimal scale kernel to the text instance with the complete shape. The ESIR~\cite{zhan2019esir} employs an innovative rectification network that corrects perspective and curvature distortions of scene texts progressively. The rectified scene text image is fed to a recognition network for better recognition performance. Chen~\etal~\cite{chen2020progressively} aims to develop an efficient and compact deep network for RGB-D salient object detection. A progressively guided alternate refinement network is proposed to refine coarse initial prediction. Saharia~\etal~\cite{saharia2021image} adapts denoising diffusion models to conditional image generation via iterative refinement, which can be used in a cascaded fashion to generate high-resolution images.
For scene text erasing task, \cite{wang2021pert} also employed the progressive strategy. Instead of using original images as supervision during erasing iterations like~\cite{wang2021pert}, we adopt the text stroke masks to provide a more accurate location. Moreover, as will be shown in the experiments, our approach outperforms previous progressive methods on popular scene text erasing benchmarks.

\section{Progressive Erasing Network}
\label{sec:Framework}

\begin{figure}[t]
    \centering
    \includegraphics[width=\linewidth]{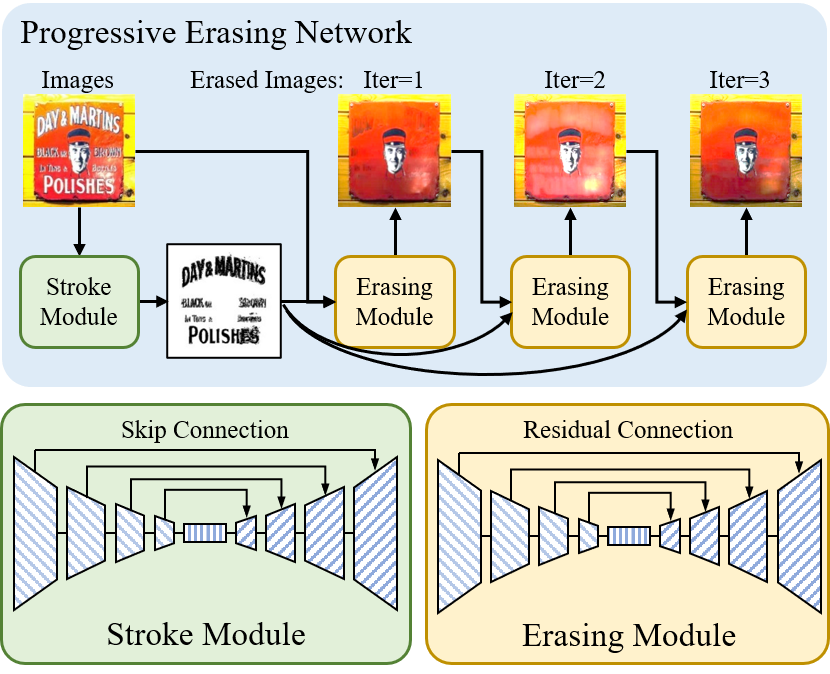}
    \caption{
    The architecture of Progressive Erasing Network. The stroke module and the erasing modules are respectively based on the encoder-decoder structure with skip connections and residual connections.
    }
    \label{fig:framework}
\end{figure}

In this section, we present the network architecture with the stroke and erasing modules. As shown in Fig.~\ref{fig:framework}, the network consists of the stroke module to compute the text stroke and the progressive erasing module in which the erasing results are being refined along with our algorithmic pipeline.

\subsection{Stroke Module}
The best case for text erasing is to preserve the background information of the scene images and fill the text regions with pixel values around the text. It is obvious that if the text stroke masks of scene images can be accurately predicted, which means that the original content can be preserved as much as possible, and then we could achieve better results according to the text location. With the input of the original scene image, we first employ ResNet~\cite{he2016deep} as the stroke encoder to extract features at each scale in the stroke module. The output feature map size of the stroke encoder is 1/16 of the input image size. Then the stroke decoder performs 4-upsampling operations to increase the spatial resolution of features consistent with the original image. The up-sampled features maps are element-wise summed with the corresponding ones from the encoding layer by the skip connection. Finally, the predicted text stroke masks are concated with the original image as input for subsequent modules. The learning target $S_{gt}$ of the stroke mask is created by subtracting the original image from ground-truth image. When the absolute value of the difference is greater than the set threshold, the pixel value corresponding to ground-truth stroke is set to 1; otherwise, the value is 0. In our implementation, the parameters of the stroke module are initialized by learning from the image pairs with data synthetic~\cite{gupta2016synthetic} and no manually labeled stroke or erased images are needed.

\begin{figure}
    \centering
    \includegraphics[width=.9\linewidth]{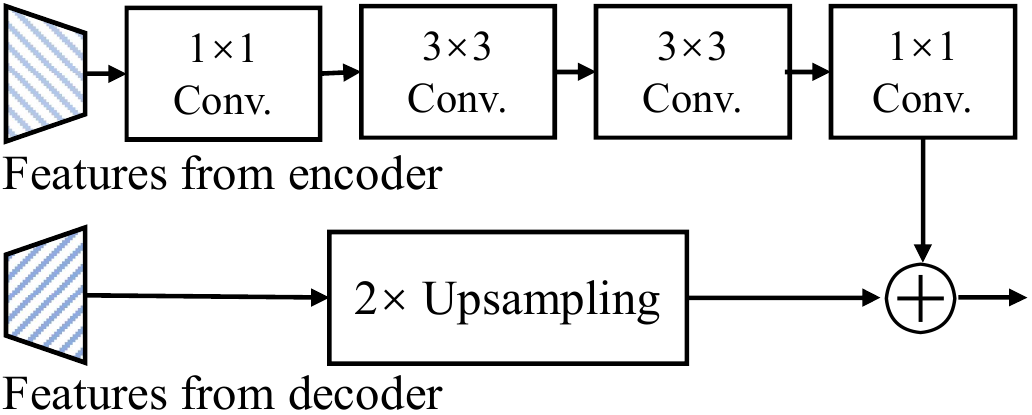}
    \caption{The residual connection used in our erasing module.}
    \label{fig:encoder-decoder}
\end{figure}

\subsection{Progressive Erasing}
The erasing module considers the image as well as the predicted stroke masks as input, as shown in Fig.~\ref{fig:framework}. The structure is also a U-Net architecture composed of an encoder and a decoder. The high-level features of erasing encoder have rich semantic information, while the low-level features have less semantic information, but more detailed information including edge, color, texture, and shape features. Here we modify the standard skip connection of the U-Net and employ the residual connection to integrate the higher-level semantics with details from the lower layers aiming to obtain effective supervisory information. As shown in Fig.~\ref{fig:encoder-decoder}, we first reduce the dimension of the low-level feature with a $1\times1$ convolution. Subsequently, two $3\times3$ convolutional layers are stacked to perform a nonlinear transformation, that can not only replace large-kernel convolutions to achieve large receptive fields, but also improve the computation efficiency. Finally, we use $1\times1$ convolution to recover the feature map channels. Besides, dilated convolutions are applied to expand the receptive field of convolution in the last layer of erasing encoder. In our method, we employ dilated convolutions with dilation rates of 2, 4, 8, and 16 respectively.

Our initial attempt shows that the single-step text erasing operation usually fails to completely remove scene text and leads to remaining rough sketches, probably due to the regions from the same image may contain texts with different shapes and distortions. Based on this observation, we decompose the erasing process into multiple iterative processes, which considerably mitigates the difficulty of each step. With the same stroke masks from the stroke module, a rough text-erased result is computed and then multiple refinements are conducted to progressively achieve optimal rectification, and we show an example with three iterations in Fig.~\ref{fig:framework}. In order to train the network effectively, we set the erasing module in different iterations with shared model weights; otherwise, the parameters will grow linearly with the number of iterations.

Our framework aims to train the Progressive Erasing Network that can cover natural scenarios. However, the learning target of erased images is difficult to achieve due to the complexity of removing text from scene images, therefore the real-world training sample is limited. Synthetic on non-text scene images is one possible solution, but the distribution inconsistency between synthetic data and real-world images still exists. In this paper, we design a pretext task by learning the self-supervised representation and weights with the stroke masks. With these parameters from the pretext task, the scene text erasing model is further trained under standard learning processes, with the losses designed to harmonize the text regions with the background context smoothly.

\begin{figure}[t]
    \centering
    \includegraphics[width=.99\linewidth]{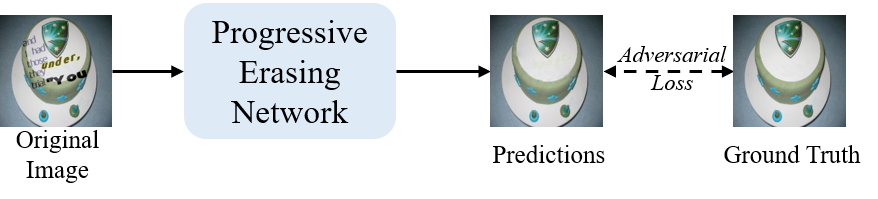}\\
    \vspace{-0.1in}
    {\footnotesize{(a) Stage-I: Learning initial parameter using synthetic dataset.}}\\
    \vspace{0.05in}
    \includegraphics[width=.99\linewidth]{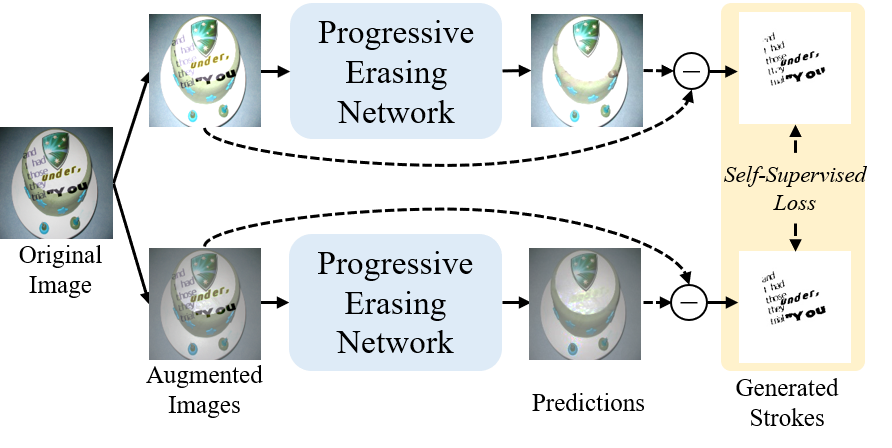}\\
    \vspace{-0.1in}
    {\footnotesize{(b) Stage-II: Learning self-supervised representation by generated strokes.}}\\
    \vspace{0.05in}
    \includegraphics[width=.99\linewidth]{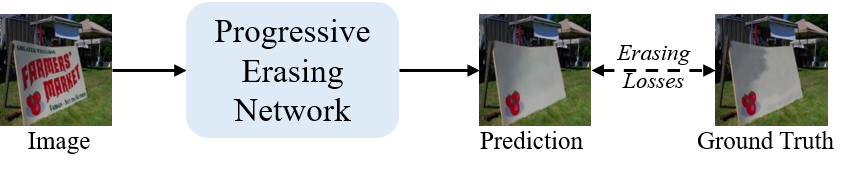}\\
    \vspace{-0.05in}
    {\footnotesize{(c) Stage-III: Fine-tuning the model with labeled data.}}
    \caption{Learning of Progressive Erasing Network. (a) Initialize parameters of progressive erasing network using synthetic data. (b) The assumption of the approach is that the text stroke masks generated by two images augmented from the same source image should be similar. $\circleddash$ is the operation of pixel-wise subtracting between the original image and erased output and finally obtaining the stroke image. (c) The proposed Progressive Erasing Network is further trained with the parameters from the pretext task as well as the labeled original and erased image pairs.}
    \label{fig:pretext}
\end{figure}

\section{Learning of Progressive Erasing Network}
\label{sec:training}
In this section, we describe the training procedure of the Progressive Erasing Network as shown in Fig.~\ref{fig:pretext}. Pretext task can be divided into two stages. In Stage-I, we first initialize parameter using synthetic dataset. In Stage-II, the erasing network further optimizes the parameter by self-supervised loss. Finally, we employ annotated datasets for scene text erasing.
\subsection{Parameter Initialization}
In order to predict reasonable stroke masks, we need to generate erase scene text properly. Therefore, in our experiment, we initialize the parameters with the adversarial loss function following~\cite{isola2017image}. Generative adversarial network consists of a generator (G) captures the data distribution, and a discriminator (D) which estimates the probability that a sample came from the training data rather than G. In our experiment, given an original image $I$ and the ground truth $I_{gt}$, PEN attempts to produce a predict image that is as real as possible by solving the following optimization problem
\begin{equation}
    \mathbb{E}_{I\sim p_{data(I)} }[\log D(I)]+\min\limits_{G} \max\limits_{D} \mathbb{E}_{I_{gt}\sim p_{I_{gt}(I_{gt})} }[\log(1-D(G(I_{gt})))]
\end{equation}
The initialize operation alternatively updates $G$ and $D$.

\subsection{Learning Self-Supervised Representation}
Large-scale labeled training samples are generally required to train the text erasing model in order to obtain better performance. However, the collection and annotation of massive datasets are time-consuming and expensive. To avoid expensive data annotations, we employ a self-supervised method to learn underlying feature representation from unlabeled scene text images without using any human annotations. Our key assumption is when augmented images are generated by some operations from the same source image, the predicted stroke masks should be relevant.

\begin{figure}[t]
    \centering
    \includegraphics[width=\linewidth]{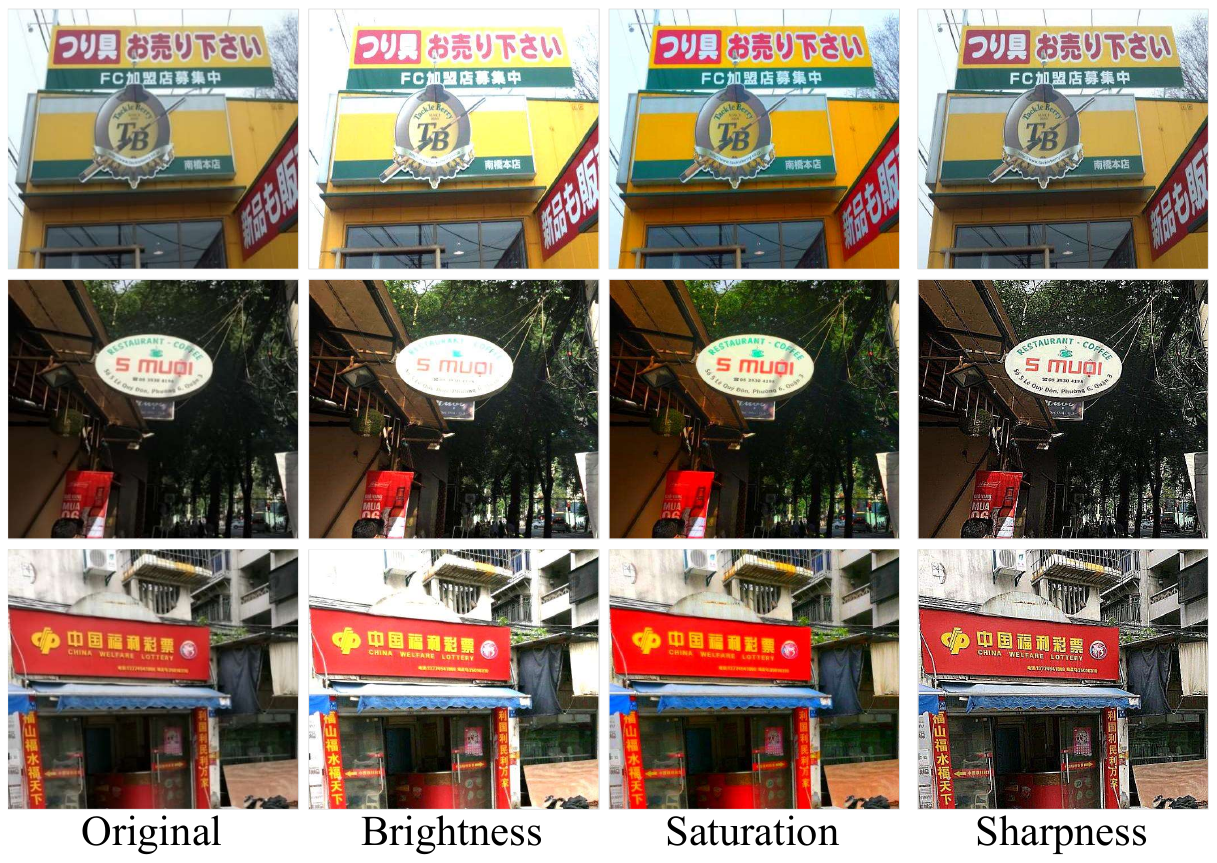}
    \caption{Different image variants employed in pretext task.}
    \label{fig:augmented_images}
\end{figure}

\vspace{0.03in}
\noindent\textbf{Augmented images} play a key role in the learning of feature representations. The complex background context will interfere final text erasing result, the image variants will improve the robustness of the feature presentation intuitively. Current stochastic augmentation schemes are mostly based on random scaling, random crop, and affine transformation. These schemes are difficult to properly meet our needs, as they may destroy the textual information of the original images. For example, if image variants are obtained from the cropped original image, it is difficult to correlate the text erasing results of the cropped variants mutually, as self-supervision is hard to learn rich patterns due to a lack of correlation information. Therefore, we randomly augment the original image with various brightness, saturation, contrast, or sharpness as shown in Fig.~\ref{fig:augmented_images}; most of these operations preserve the text information of the original image. As validated by our experiments, the generated image variants provide important clues for the subsequent learning procedures.

\vspace{0.03in}
\noindent\textbf{Self-supervised loss.}
To learn the self-supervised representation, any scene text images can be used without the annotation of text locations or text contents. With going through the Progressive Erasing Network, the predictions for erased images are generated, and the strokes are computed with the same step in the stroke module, \ie, subtracting the erased images from the augmented images. To keep the robustness of the image representation in the pretext task, we aim to ensure the consistency of computed stroke masks from the two streams. Given a set of $N$ training samples $\{ {{I_{i}}}\}$, two augmented batches of data are denoted as $\{I_{i,1}\}$ and $\{I_{i,2}\}$, respectively. With the Progressive Erasing Network $f(\cdot)$, the stroke mask $S_{i,j}$ is computed by the operation of pixel-wise subtracting between $I_{i,j}$ and $f(I_{i,j})$. Our loss function for the self-supervision takes the form of a simple $L_1$ loss defined as:
\begin{equation}
    L_{ss}= \frac{1}{N}\sum_{i = 1}^N L_{1}(S_{i,1},S_{i,2})
\end{equation}

\subsection{Training PEN with Annotated Datasets}
As shown in Fig.~\ref{fig:pretext}(c), the training of the text erasing model follows a supervised pipeline with standard benchmark settings, and the parameters of the PEN are set with the pretext task. Inspired by the previous work~\cite{tang2021stroke,liu2020erasenet,tursun2020mtrnet++}, we utilize four loss terms to measure the structural and textural differences between the output image and ground truth, including reconstruction loss, content loss, style loss, and adversarial loss of two-stream discriminator.

\vspace{0.03in}
\noindent\textbf{Reconstruction loss} aims to guide pixel-level reconstruction between the output image ${I'}$ and the ground truth image ${I_{\rm gt}}$ assist with the predicted stroke mask ${S}$, where more weight is added to the text stroke regions. The reconstruction loss is defined as:
\begin{equation}
    {L_{\rm r}} = {{\lambda _{\rm r1}}{{\| {{ S}\odot( {{I'} - { I_{\rm gt}}} )} \|}_1}} + {\lambda _{\rm r2}}\| {({1 - { S}} )\odot( {{I'}-{ I_{\rm gt}}} )} \|_1
\end{equation}

\vspace{0.03in}
\noindent\textbf{Content loss} is introduced to penalize the discrepancy between different high-level features of $I'$, the corresponding ground truth $I_{\rm gt}$, and composed image $I''$ on certain layers in the CNN. $I''$ is the output image with the non-text regions being set to the ground truth. We extract the features through a VGG-16 network pre-trained on ImageNet and enforce matching the features of the corresponding layers, which will facilitate the network in detecting and erasing text regions. The content loss is defined as follows:
\begin{equation}
    {L_{\rm c}} = \sum\limits_{ n} [{\| {{\phi _{ n}}(I') - {\phi _{n}}(I_{\rm gt})}\|}_1 +  {{{\| {{\phi _{n}}( {{I''}} ) - {\phi _{n}}( {{I_{\rm gt}}} )} \|}_1}}]
\end{equation}

\begin{equation}
    I'' = {I_{\rm gt}} \odot ( {1 - {S}} ) + I' \odot {S}
\end{equation}
where $\phi_{\rm n}$ denotes the feature map of the $n-$th pooling layers of the pre-trained VGG-16.

\vspace{0.03in}
\noindent\textbf{Style loss} is motivated by the recent success of neural style transfer~\cite{gatys2015neural} to penalize the differences in image style, such as color, texture, and pattern. The style loss is defined follows~\cite{liu2020erasenet}:
\begin{equation}
    {L_{\rm s}} = \sum\limits_{n} {\frac{1}{{{h_{n}}{w_{n}}{c_{n}}}}\| {{\phi _{n}}{{( {{I'}} )}^{T}}{\phi _{n}}( {{I'}} ) - {\phi _{n}}{{( {{I_{\rm gt}}} )}^{T}}{\phi _{n}}( {{I_{\rm gt}}} )} \|_1}
\end{equation}
where $(h_{n},{w_{n}},{c_{n}})$ is the shape of the feature map, $\phi_{n}$ follows the same setting as the content loss.

\begin{figure}[t]
    \centering
    \includegraphics[width=.9\linewidth]{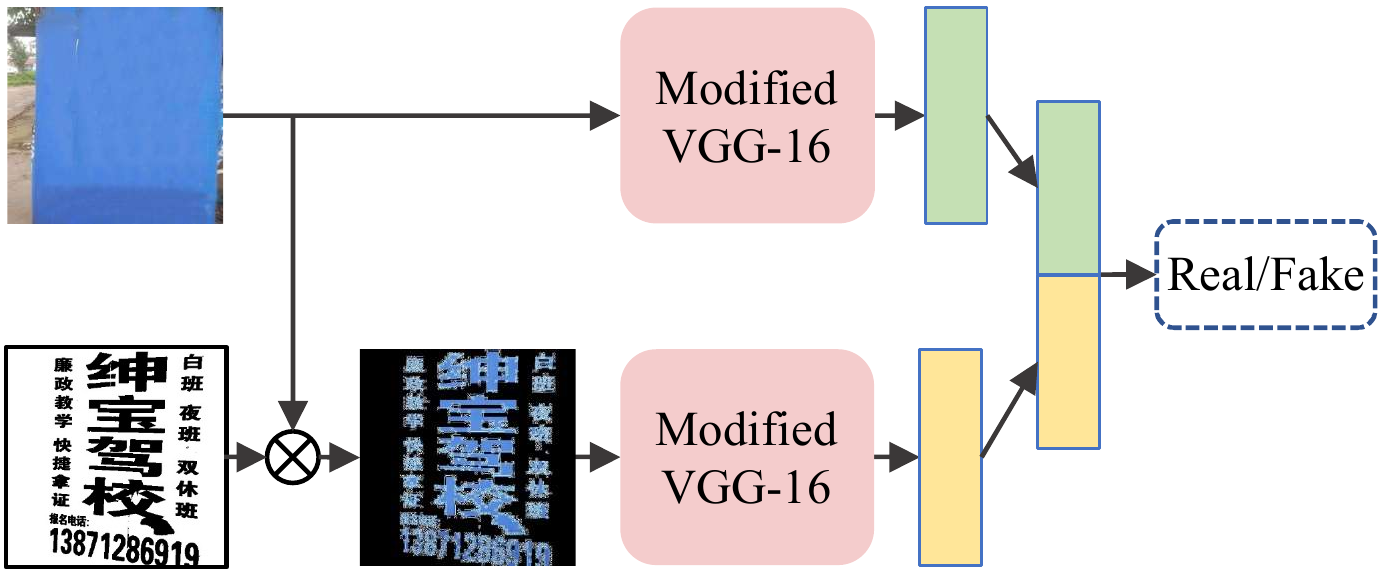}
    \caption{Two-stream discriminator for model training.}
    \label{fig:discriminator}
\end{figure}

\vspace{0.03in}
\noindent\textbf{Adversarial loss from two-stream discriminator.}
Inspired by~\cite{iizuka2017globally,yu2019free,miyato2018spectral,liu2020erasenet}, we employ a two-stream discriminator $D'$ to distinguish genuine images from the generated ones by estimating the feature statistics of global and local output (Fig~\ref{fig:discriminator}). We drop the fully connected layer of VGG-16 and modify the convolution layers with the kernel size of 4 and stride of 2. Leaky ReLU is used with a slope of 0.2 after each layer. The local branch shares the same pattern as the global stream, where the input is text regions inpainted with the background content. Finally, we calculate the adversarial loss $L_a$ to generate realistic erased results. The $L_a$ is defined as:
\begin{equation}
    L_a=-\mathbb{E}[D_{ts}(I',S_{gt})]
\end{equation}
We train the discriminator $D_{ts}$ iteratively together with the progressive erasing network with a loss function $L_D=\mathbb{E}[ReLU(1-D_{ts}(I_{gt}, S_{gt}))] + \mathbb{E}[ReLU(1+D_{ts}(I', S_{gt}))]$.

Finally, the overall erasing loss are summarized as:
\begin{equation}
    L = L_r + \lambda_cL_c + \lambda_sL_s + \lambda_aL_a
\end{equation}
where $\lambda_c$, $\lambda_s$ and $\lambda_a$ are the trade-off parameters.

\begin{figure}[t]
  \centering
  \includegraphics[width=.99\linewidth]{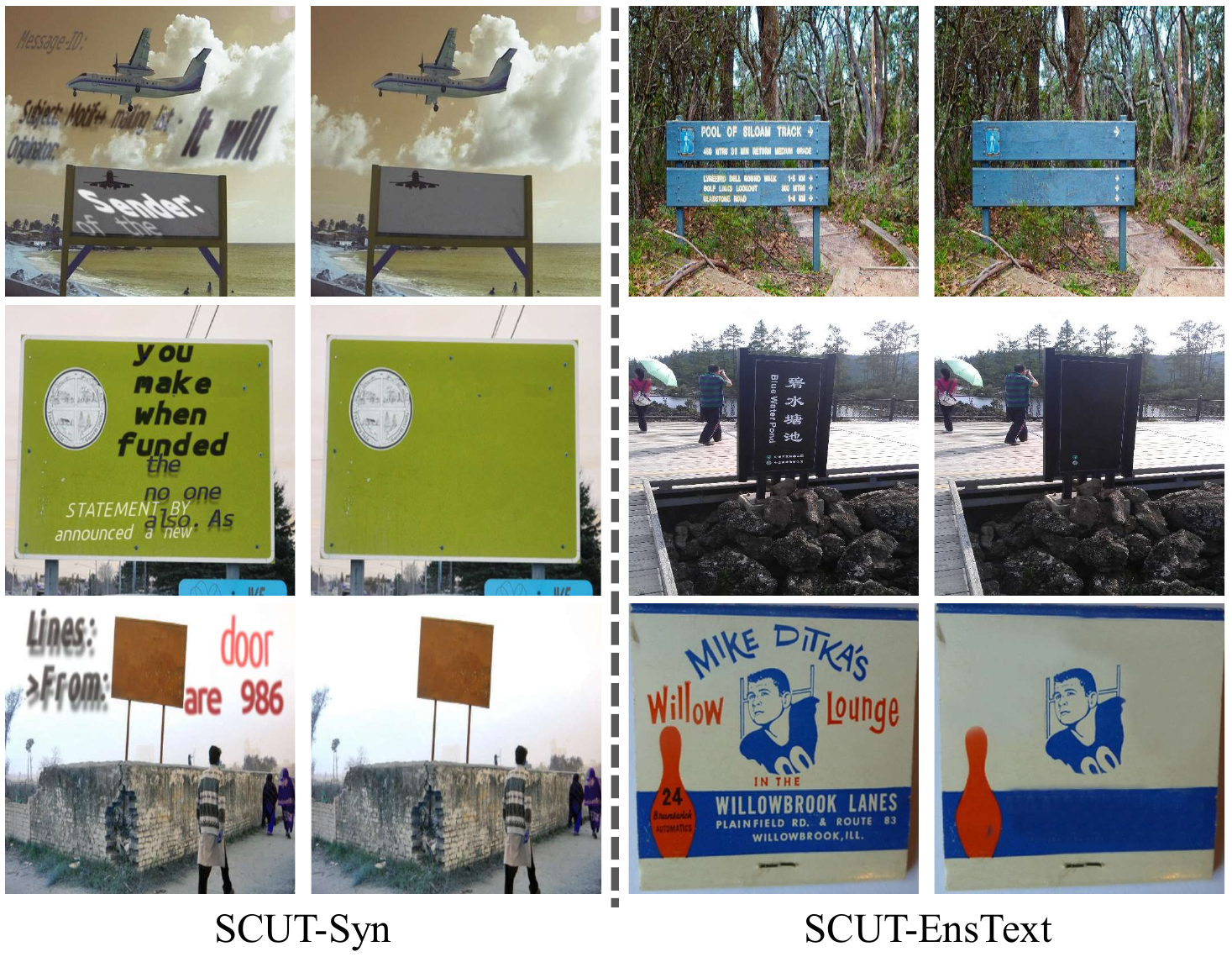}
  \caption{Example scene text images and the corresponding erased images of the synthetic and real-world datasets.}
  \label{fig:labeleddataset}
\end{figure}

\begin{table*}[t]
  \caption{Comparison with state-of-the-art methods on SCUT-Syn dataset. Bold text denotes the top result, while underlined text corresponds to the runner-up.}
  \label{tab:quantitative_SCUT-Syn}
  \centering
  \small
  \begin{tabular}{c|cccccc|ccc}
    \hline
    Method & PSNR$\uparrow$ & SSIM$\uparrow$ & MSE$\downarrow$ & AGE$\downarrow$ & pEPS$\downarrow$ & pCEPS$\downarrow$ & Precision$\downarrow$ & Recall$\downarrow$ & F1$\downarrow$ \\ \hline
    Original Images & - & - & - & - & - & - & 76.4 & 44.3 & 56.1 \\
    Scene text eraser~\cite{nakamura2017scene} & 14.68 & 46.13 & 0.7148 & 13.29 & 0.1859 & 0.0936 & - & - & - \\
    Pix2Pix~\cite{isola2017image} & 25.60 & 89.86 & 0.2465 & 5.60 & 0.0571 & 0.0423 & 70.0 & 29.3 & 41.4 \\
    EnsNet~\cite{zhang2019ensnet} & 37.36 & 96.44 & 0.0021 & 1.73 & 0.0069 & 0.0020 & 57.3 & 14.3 & 22.9 \\
    EraseNet~\cite{liu2020erasenet} & 38.32 & 97.67 & \textbf{0.0002} & 1.60 & 0.0048 & \textbf{0.0004} & - & - & - \\
    MTRNet~\cite{tursun2019mtrnet} & 29.71 & 94.43 & 0.0004 & - & -  & - & 35.8 & \textbf{0.3} & \textbf{0.6} \\
    MTRNet++~\cite{tursun2020mtrnet++} & 34.55 & \textbf{98.45} & 0.0004 & - & - & - & 50.4 & 1.4 & 2.6 \\
    PERT~\cite{wang2021pert} & \textbf{39.40} & 97.87 & \textbf{0.0002} & 1.41 & 0.0045 & 0.0006 & - & - & - \\ \hline
    PEN & \underline{39.26} & \underline{98.03} & \textbf{0.0002} & \textbf{1.29} & \textbf{0.0038} & \textbf{0.0004} & \textbf{16.7} & \textbf{0.3} & \textbf{0.6} \\
    PEN* & 38.87 & 97.83 & 0.0003 & \underline{1.38} & \underline{0.0041} & \textbf{0.0004} & \underline{23.5} & 0.7 & 1.4 \\ \hline
    \end{tabular}
\end{table*}

\begin{table*}[t]
  \caption{Comparison with state-of-the-art methods on SCUT-EnsText dataset. $\dagger$As reported in~\cite{tang2021stroke}, we consider the images as the inputs of the comparing methods, and the text bounding boxes are obtained by the CRAFT detector~\cite{baek2019character}.}
  \label{tab:quantitative_SCUT-EnsText}
  \centering
  \small
  \begin{tabular}{c|cccccc|ccc}
    \hline
    Method & PSNR$\uparrow$ & SSIM$\uparrow$ & MSE$\downarrow$ & AGE$\downarrow$ & pEPS$\downarrow$ & pCEPS$\downarrow$ & Precision$\downarrow$ & Recall$\downarrow$ & F1$\downarrow$ \\ \hline
    Original Images & - & - & - & - & - & - & 79.4 & 69.5 & 74.1 \\
    Scene text eraser~\cite{nakamura2017scene} & 25.47 & 90.14 & 0.0047 & 6.01 & 0.0533 & 0.0296 & 40.9 & 5.9 & 10.2 \\
    Pix2Pix~\cite{isola2017image} & 26.70 & 88.56 & 0.0037 & 6.10 & 0.0480 & 0.0270 & 69.7 & 35.4 & 47.0 \\
    EnsNet~\cite{zhang2019ensnet} & 29.54 & 92.47 & 0.0024 & 4.16 & 0.0307 & 0.0136 & 68.7 & 32.8 & 44.4 \\
    EraseNet~\cite{liu2020erasenet} & 32.30 & 95.42 & 0.0015 & 3.02 & 0.0160 & 0.0090 & 54.1 & 8.0 & 14.0 \\
    PERT~\cite{wang2021pert} & 33.25 & \textbf{96.95} & 0.0014 & 2.18 & 0.0136 & 0.0088 & 52.7 & 2.9 & 5.4 \\
    Stroke-based method~\cite{tang2021stroke}$\dagger$ & \underline{35.34} & 96.24 & 0.0009 & - & - & - & - & - & - \\ \hline
    PEN & 35.21 & 96.32 & \underline{0.0008} & \underline{2.14} & \underline{0.0097} & \underline{0.0037} & \underline{33.5} & \underline{2.6} & \underline{4.8} \\
    PEN* & \textbf{35.72} & \underline{96.68} &\textbf{0.0005} & \textbf{1.95} & \textbf{0.0071} &\textbf{0.0020} & \textbf{26.2} & \textbf{2.1} & \textbf{3.9} \\ \hline
    \end{tabular}
    \end{table*}

\begin{figure*}[t]
  \centering
  \includegraphics[width=.99\linewidth]{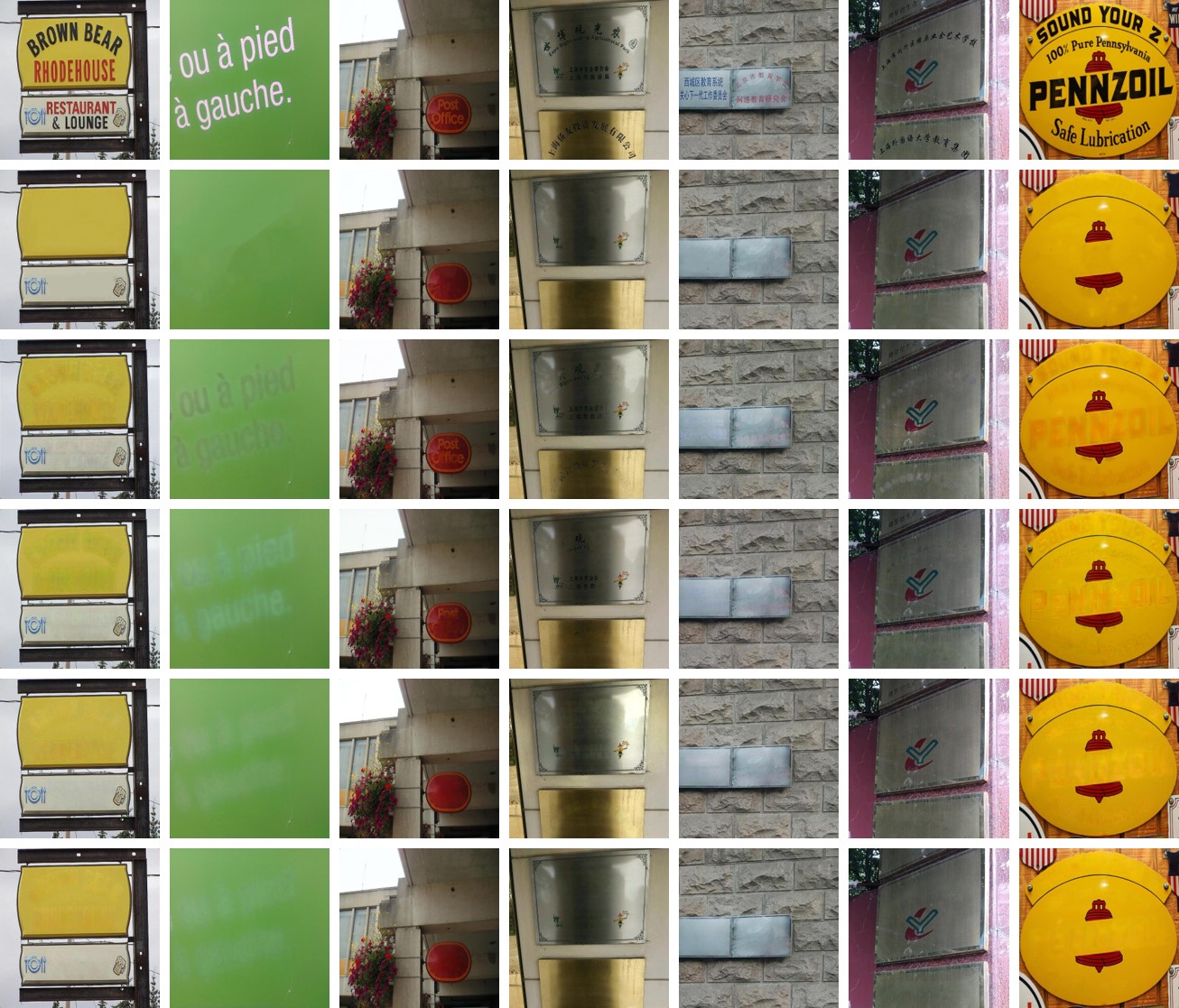}
  \caption{Qualitative results of the proposed approach and previous methods examples on SCUT-EnsText. From top to buttom: original images, ground-truth erased images (GT), results by Pix2Pix~\cite{isola2017image}, EraseNet~\cite{liu2020erasenet}, Stroke-based erasing~\cite{tang2021stroke}, and the proposed PEN*.}
  \label{fig:stoa}
\end{figure*}

\section{Experiments}
\label{sec:exp}
In this section, we first introduce the implementation details, settings, datasets, and metrics for our evaluations. Then, we experimentally compare the performance of our method on the benchmarks. Finally, we conduct ablation studies to further investigate the functionality of each module.

\subsection{Datasets for Evaluation}
We evaluate the performance of the proposed framework with both real-world (SCUT-EnsText) and synthesis (SCUT-Syn) image datasets~\cite{liu2020erasenet}. Fig.~\ref{fig:labeleddataset} illustrates some sample image pairs.

\begin{itemize}
  \item SCUT-EnsText contains 3,562 images with diverse text characteristics, including text shape (horizontal text, arbitrary quadrilateral text, and curved text) and languages(English and Chinese). Following previous works, we randomly select approximately 70\% of the images for training and the rest for testing.
  \item SCUT-Syn contains 8,000 images for training and 800 images for testing, both created by synthesis technology. The background images were collected from ICDAR 2013 and ICDAR MLT-2017, and the text instances in the background images were manually erased. We evaluated our method using only test images.
\end{itemize}

\subsection{Implementation Details}
There are several stages to train our network. We first train stroke module by synthetic dataset~\cite{gupta2016synthetic}. Then we froze the stroke module and use self-supervised learning (pretext task) to further train the erasing network. The pretext task can be divided into two stages as shown in Figure~\ref{fig:pretext} (Fig. 4 in the revised manuscript). In Stage-I, we train progressive erasing network using synthetic data. In Stage-II, we assume that the text strokes generated by the augmented images are similar to further train the erasing network. Finally, the stroke module and erasing module are jointly trained on annotated datasets (Stage-III) by the overall erasing loss as described in Section \ref{sec:training}. Here we empirically set $\lambda _{\rm r1}=10$, $\lambda _{\rm r2}=2$, $\lambda_{\rm c}=0.1$, $\lambda_{\rm s}=150$, and $\lambda_{\rm G}=0.1$.

The pretext task is an important prerequisite for our erasing model. We obtain the initial weight of the network by existing scene text images. We design two datasets for the pretext task. The first set is the training images of SCUT-Syn, which are generated by rendering the text on the various background images. We denote our approach as {\bf PEN} ({\bf P}rogressive {\bf E}rasing {\bf N}etwork).
Besides, in order to evaluate the impact of the real-world text images, we also collect scene text images from the existing scene text detection datasets: CTW1500~\cite{liu2019curved}, ICDAR17-RCTW~\cite{shi2017icdar2017}, Total-Text~\cite{ch2017total}, and MSRA-TD500~\cite{yao2012detecting}. We collected a total of 10589 training images, including 8034 ICDAR17-RCTW images, 1000 CTW1500 images, 1255 Total-Text images and 300 MSRA-TD500 images. Note that although these datasets are with the annotated text locations, only the text images themselves are used in the paper for the pretext task, as the only precondition is the image contains text strokes. We denote the model trained with extra scene text images as {\bf PEN*} in the evaluation.

The proposed method is implemented using Pytorch~\cite{paszke2019pytorch}. Original images are resized to the size $512 \times 512$ before feeding the network with random horizontal flip and rotation. Our model is trained by an Adam optimizer~\cite{kingma2014adam} with learning rate $10^{-4}$ and parameters ${\beta _1}=0.5$, ${\beta _2}=0.9$. For the discriminator subnet, learning rate is set to $10^{-5}$ and $\beta=(0.0, 0.9)$. 

\subsection{Evaluation Metrics}
Image inpainting metrics are used to evaluate the quality or realistic degree of the results, \ie, peak signal-to-noise ratio (PSNR), structural similarity index (SSIM), mean squared error (MSE), average of the gray-level absolute (AGE), percentage of error pixels (pEPs), and percentage of clustered error pixels (pCEPS). Higher PSNR, SSIM, and lower MSE, AGE, pEPs, pCEPS values indicate better-erased quality.

We also follow the evaluation protocol with text detection to quantify text erasure ability~\cite{nakamura2017scene}. For an erased image, a state-of-the-art text detector~\cite{baek2019character} is applied to have the detection results and the performances are measured with precision, recall, and F1. Lower precision and recall indicate that less text was detected and more text is erased by the model.

\subsection{Comparison with State-of-the-Arts}
We compare our method on both real-world and synthesis testing sets against the recent approaches, \ie,
\begin{itemize}
  \item {Scene text eraser~\cite{nakamura2017scene}} is the first method to address the scene text erasing task which processes image patches cropped by sliding windows.
  \item {Pix2Pix~\cite{isola2017image}} is a general-purpose solution for image-to-image translation problems.
  \item {EnsNet~\cite{zhang2019ensnet}} utilizes an end-to-end network to effectively erase scene texts from whole images. 
  \item {MTRNet~\cite{tursun2019mtrnet}} and {MTRNet++~\cite{tursun2020mtrnet++}} applies a mask-based text erasing network.
  \item {EraseNet~\cite{liu2020erasenet}} consists of a coarse-erasure and a refinement sub-network to erase texts.
  \item {PERT~\cite{wang2021pert}} also uses progressive strategy for accurate and exhaustive text erasing.
  \item {Stroke-based erasing~\cite{tang2021stroke}} is also based on stroke; it operates on the cropped text images from human labeling or auxiliary text detector. 
\end{itemize}

The experimental results of our method compared with those of the state-of-the-art approaches on the synthesis \textbf{SCUT-Syn} dataset are given in Table~\ref{tab:quantitative_SCUT-Syn}. PEN shows very competitive performance with the best results on four image inpainting metrics (MSE, AGE, pEPS, and pCEPS) and all the text detection evaluation metrics. We see that PEN outperforms all the early erasing approaches (Scene text eraser and EnsNet) as well as the general-purpose baseline (Pix2Pix). Compared with PERT~\cite{wang2021pert} which also applies progressive strategy, we achieve better performance on four image inpainting metrics, while the performance of the other two metrics reaches the same magnitude. The SSIM of our method is slightly higher than that of MTRNet++~\cite{tursun2020mtrnet++}, as they adopt a larger synthetic dataset ~\cite{gupta2016synthetic} with around 800,000 images. We also study the performance by learning self-supervision from real-world images and then applying it for synthesis; the results of PEN* are not as good as PEN, indicating that data domain consistency for the pretext task is important for learning the feature representations.

\begin{figure}[t]
  \centering
\includegraphics[width=\linewidth]{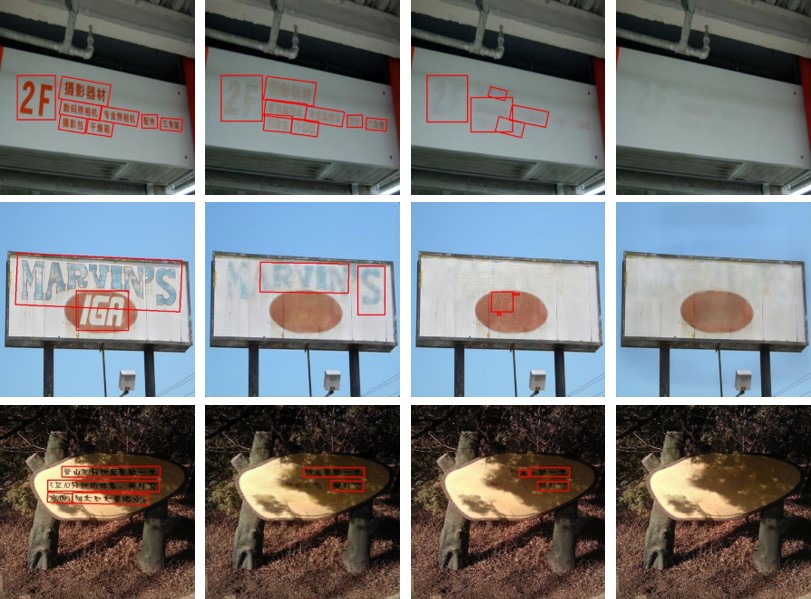}
\caption{The effect of detecting text in images processed by text erasing. From left to right: detection results of original images, erased images by~\cite{isola2017image},~\cite{liu2020erasenet}, and the proposed PEN*.}
\label{fig:detection}
\end{figure}

We now report results on the real-world \textbf{SCUT-EnsNet} dataset, as summarized in Table~\ref{tab:quantitative_SCUT-EnsText}. PEN* achieves the best performance in most evaluation metrics. We present the visual effort for the erased images in Fig.~\ref{fig:stoa}. The erasing qualities of PEN* are notably better than those comparing approaches (Pix2Pix, EraseNet, and stroke-based erasing). Compared with PERT~\cite{wang2021pert}, the SSIM of ours is lower (0.27) than the SSIM of PERT, while the PSNR is higher (2.47dB). The results of PEN for the real-world dataset are lower than PEN*, but still better than those from previous approaches. This highlights the advantages of using self-supervised representations to improve the erasing ability. For the text detection evaluation, most text has been successfully erased and some examples of text detection results are displayed in Fig.~\ref{fig:detection}.

\subsection{Ablation Study}
We conduct some comparison experiments to verify the benefits of various aspects of the model including training manner, progressive strategy, and the stroke module. The ablation study is performed with the default setting of PEN* on the SCUT-EnsText dataset.

\begin{figure}[t]
  \centering
  \includegraphics[width=.8\linewidth]{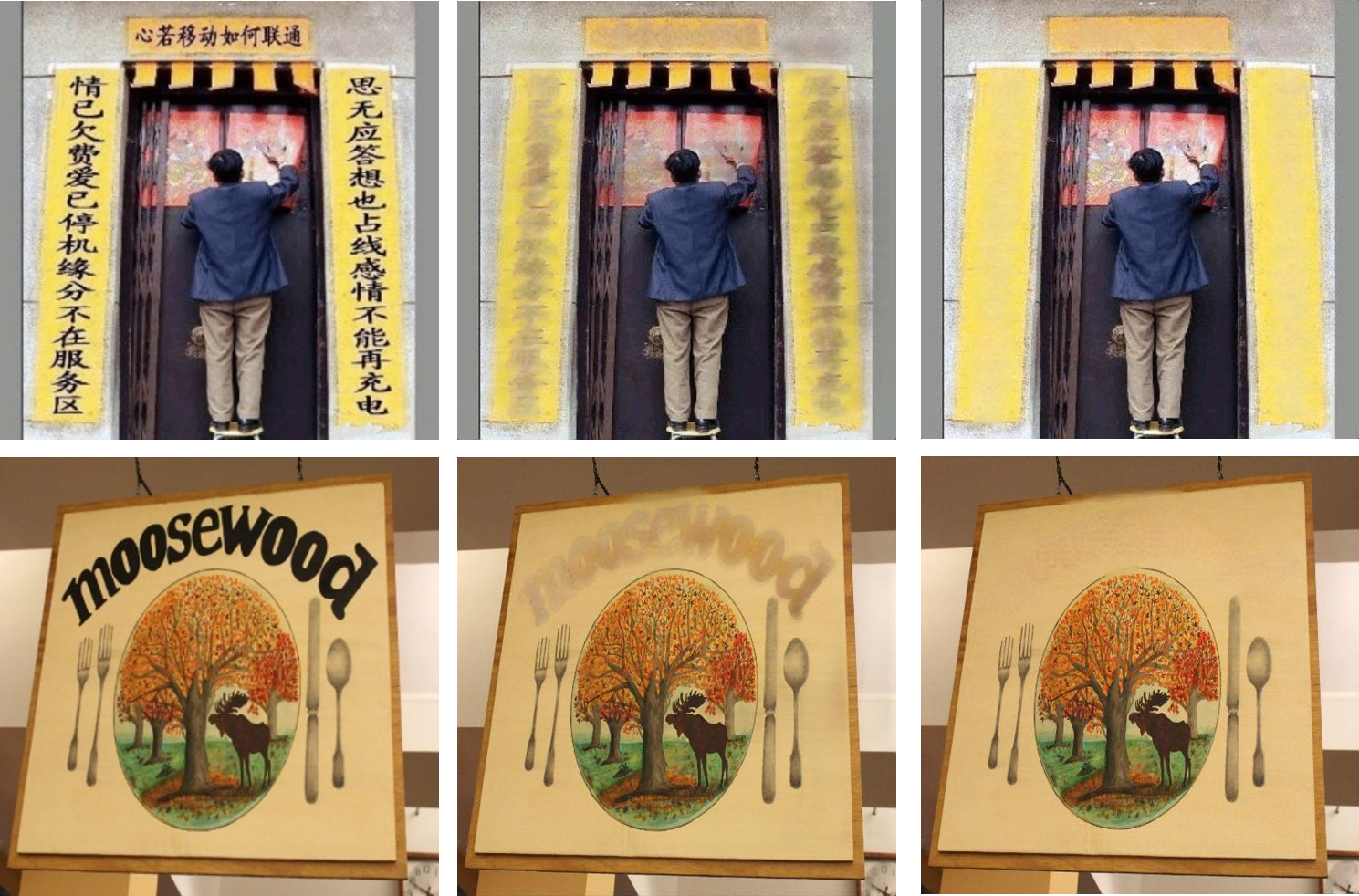}\\
  \vspace{-0.05in}
  {\footnotesize (a) The effect of fine-tuning. From left to right: input images, results with fine-tuning, and results without fine-tuning.}\\
  \vspace{0.05in}
  \includegraphics[width=.8\linewidth]{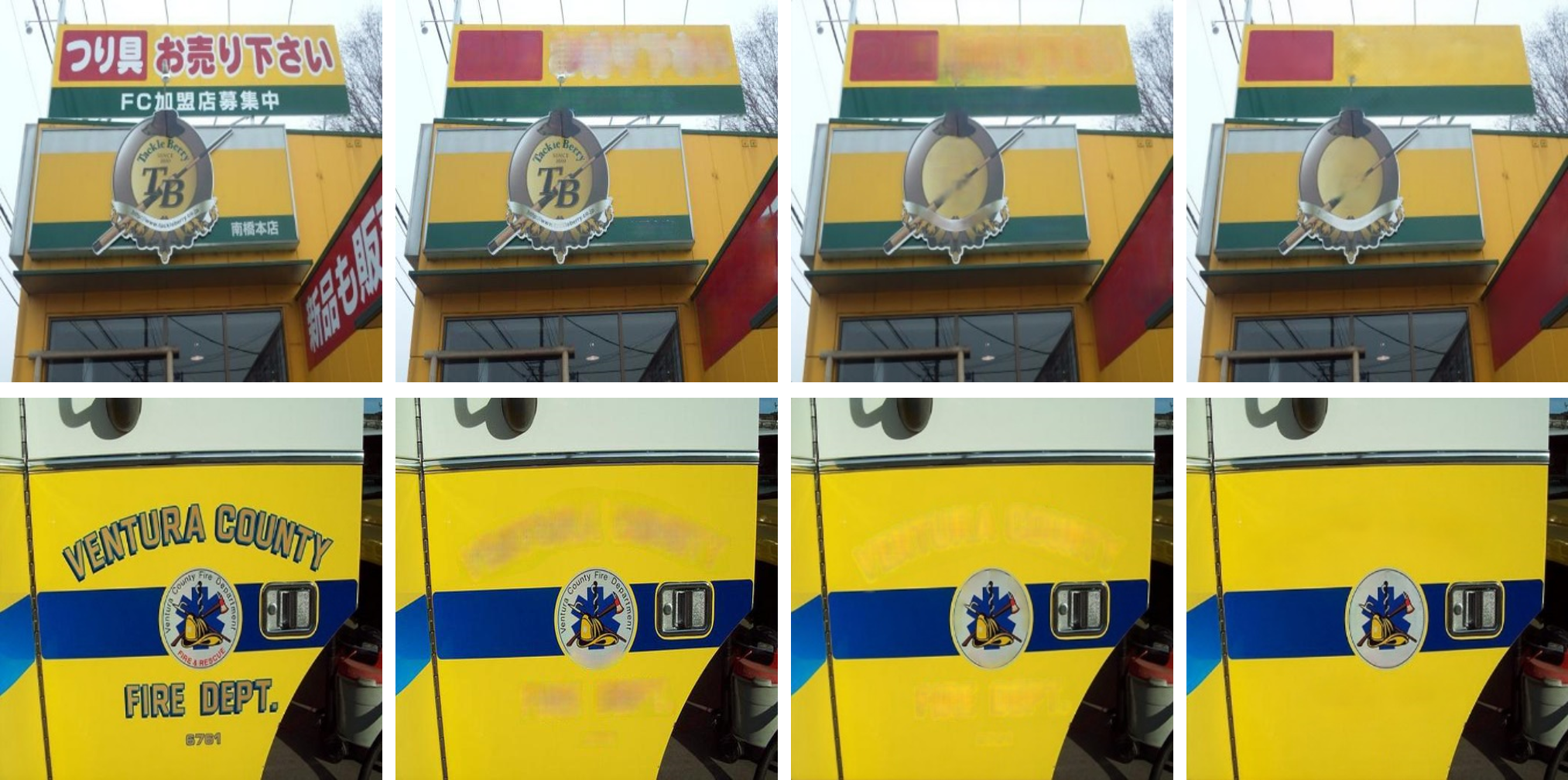}\\
  \vspace{-0.05in}
  {\footnotesize (b) The effect of self-supervision. From left to right: input images, erasing results without self-supervision (setting(b) in Table 1), result with self-supervision from synthetic (setting(d)) or real world images (setting(e)).}\\
  \vspace{0.05in}
  \includegraphics[width=.8\linewidth]{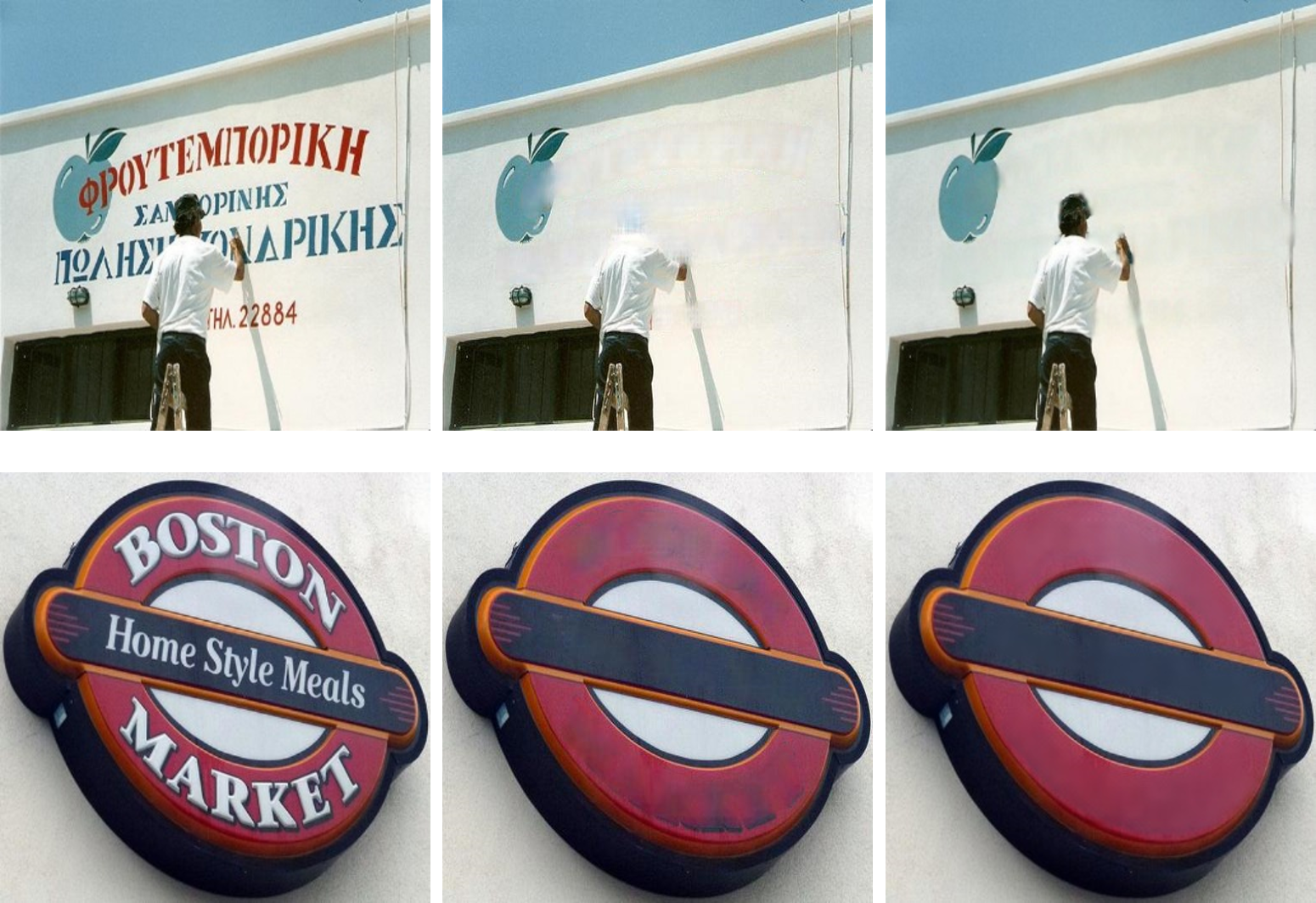}\\
  \vspace{-0.05in}
  {\footnotesize (c) Comparison of traditional supervised training and the proposed manner.}\\
  \caption{Ablation study on the training manner.}
  \label{fig:fig10new}
\end{figure}

\vspace{0.03in}
\noindent\textbf{Training manner.} We set up different combinations of various pretext task and fine-tuning to verify the impact of the training manner. The pretext task is designed to enhance the feature representation of unlabeled text images. Table~\ref{tab:design} shows the ablation investigation on the effects of training manner. Here setting (a) employs the erasing network only supervised by Stage-I and Stage-II; (b) trains the network without Stage-II; (c) traditional supervised training by using synthetic data and real data for fine-tuning with the erasing losses in Stage-III; (d) trains the network with the whole stages by using synthetic images in Stage-II; (e) replaces the synthesis in (d) with real images.

Directly applying the network for pretext reaches a low performance (PSNR=22.86, SSIM=80.04). As shown in Fig.~\ref{fig:fig10new}(a), we can observe that it can act on the text regions, but the erasing effects are limited comparing with the results by the full model. Incorporating the self-supervision in the pretext task and the model training with training sets can further boost the erasing results (setting (d) and (e)) and achieve more natural erasing results (see Fig.~\ref{fig:fig10new}(b)), which demonstrates that adding self-supervision helps the network capture more discriminative features. We also observe pretext task on unlabeled real-world images can achieve better results compared with synthetic images, probably due to their similar visual patterns in the text regions. We also compare the proposed self-supervised manner with traditional supervised learning (setting (c)). As shown in in Fig.~\ref{fig:fig10new}(c), we observe that incorporating the self-supervision with unlabeled real data outperforms the supervised approach using synthetic data with groundtruth.

\begin{table}[t]
  \caption{The performance of different settings for model training. See Section 5.5 for more details.}
  \label{tab:design}
  \centering
  \small
  \begin{tabular}{c|cccccc}
    \hline
    Settings & PSNR$\uparrow$ & SSIM$\uparrow$ & MSE$\downarrow$ & AGE$\downarrow$ & pEPS$\downarrow$ \\ \hline
    (a) & 22.86 & 80.04 & 0.0157 & 9.93 & 0.1018 \\
    (b) & 34.22 & 96.17 & 0.0012 & \underline{2.03} & 0.0125 \\
    (c) & {35.05} & {96.27} & \underline{0.0008} & {2.33} & {0.0113} \\
    (d) & \underline{35.21} & \underline{96.32} & \underline{0.0008} & 2.14 & \underline{0.0097} \\
    (e) & \textbf{35.72} & \textbf{96.68} & \textbf{0.0005} & \textbf{1.95} & \textbf{0.0071} \\ \hline
  \end{tabular}
\end{table}

\begin{figure}[t]
  \centering
  \includegraphics[width=.99\linewidth]{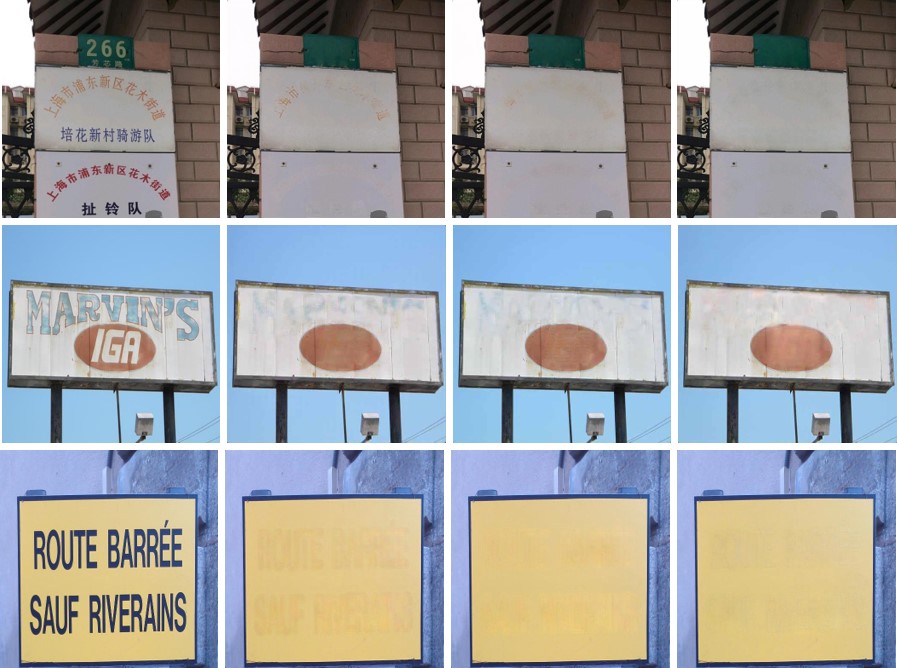}
  \caption{Examples with different iterations numbers. From left to right: inputs, results with \#iter=1,2,3.}
  \label{fig:fig10b}
\end{figure}

\vspace{0.03in}
\noindent\textbf{Progressive strategy.} We also investigate the effect of the proposed progressive strategy, which is also the number of erasing iterations. We conduct experiments with 1, 2, 3, 4 and 6 iteration numbers. As shown in Table~\ref{tab:design2}, the quality of generated images increases continuously when the number of iterations increases.
We can see that adding more iteration numbers further boosts the performance until 3 (PSNR=35.72), after which the performance tends to be saturate. 
On the other hand, the inference time has nearly linear relationship with iteration number. 
Therefore, we choose iteration of 3 to trade off performance and speed. We also show some qualitative examples in Fig.~\ref{fig:fig10b}. The residual text of intermediate results is gradually removed, thereby justifying the importance of the progressive strategy.

\begin{table}[t]
  \caption{The performance and inference time of different number of iterations on SCUT-EnsText. The inference time is measured in millisecond with ten different $512\times512$ images as input.}
  \label{tab:design2}
  \centering
  \small
  \begin{tabular}{c|cccccc}
    \hline
    \#iter. & PSNR$\uparrow$ & SSIM$\uparrow$ & MSE$\downarrow$ & AGE$\downarrow$ & pEPS$\downarrow$ & Time$\downarrow$ \\ \hline
    1 & 34.53 & 95.97 & 0.0007 & 2.86 & 0.0113 & 6.71$\pm$0.73  \\
    2 & 35.47 & 96.31 & 0.0006 & 2.31 & 0.0081 & 13.62$\pm$0.88 \\
    3 & 35.72 & 96.68 & 0.0005 & 1.95 & 0.0071 & 19.74$\pm$0.92 \\ 
    4 & 35.77 & 96.73 & 0.0005 & 1.86 & 0.0066 & 27.03$\pm$1.37 \\ 
    6 & 35.80 & 96.76 & 0.0005 & 1.84 & 0.0065 & 40.01$\pm$2.54 \\ 
    \hline
  \end{tabular}
\end{table}

\begin{figure}[t]
  \centering
  \includegraphics[width=.99\linewidth]{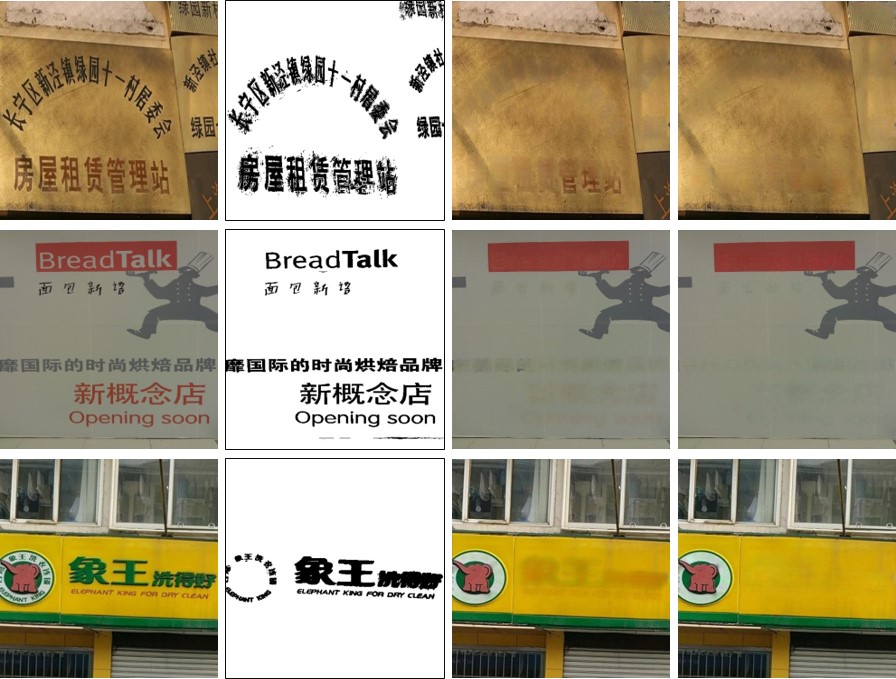}
  \caption{Erasing results with the stroke module. From left to right: input images, the predicted text stroke masks, results without/with stroke module.}
  \label{fig:fig10c}
\end{figure}

\begin{table}[t]
  \caption{The effect of the stroke module.}
  \label{tab:design3}
  \centering
  \small
  \begin{tabular}{c|cccccc}
    \hline
    Design & PSNR$\uparrow$ & SSIM$\uparrow$ & MSE$\downarrow$ & AGE$\downarrow$ & pEPS$\downarrow$ \\ \hline
    w/o stroke & 35.17 & 96.22 & 0.0006 & 2.63 & 0.0073 \\
    w. stroke & \textbf{35.72} & \textbf{96.68} & \textbf{0.0005} & \textbf{1.95} & \textbf{0.0071} \\ \hline
  \end{tabular}
\end{table}

\vspace{0.03in}
\noindent\textbf{Stroke module.} Text strokes provide important positional information for the text erasing task, which assist to distinguish scene texts from background image accurately. We can keep background information as much as possible according to the location information. Besides, the stroke module is plugin component for some existing erasing methods. As shown in Table~\ref{tab:design3}, when inserting the stroke module into the framework, the performance of the model is improved. Some examples are illustrated in Fig.~\ref{fig:fig10c}. With the guidance of the text stroke mask, we can see that the texts are removed clearly from the original scene images.

\section{Conclusions}
\label{sec:conclusion}
In this paper, we introduced a novel framework for scene text erasing, which consists of a pretext task to learn the self-supervised representation, as well as model training with original and erased scene image pairs. The Progressive Erasing Network is proposed with a stroke module that considers the text stroke, and an erasing module that removes text progressively. During the pretext task, unlabeled text images were utilized to constrain the entire network. Then the annotated training samples were employed to further optimize the model combined with the erasing loss. Experimental comparisons with the state-of-the-art approaches on SCUT-EnsText and SCUT-Syn showed the effectiveness and efficiency of our proposed self-supervision and progressive erasing for the text erasing task.

\ifCLASSOPTIONcaptionsoff
  \newpage
\fi

\bibliographystyle{IEEEtran}
\bibliography{total}

\begin{thebibliography}{10}
\providecommand{\url}[1]{#1}
\csname url@samestyle\endcsname
\providecommand{\newblock}{\relax}
\providecommand{\bibinfo}[2]{#2}
\providecommand{\BIBentrySTDinterwordspacing}{\spaceskip=0pt\relax}
\providecommand{\BIBentryALTinterwordstretchfactor}{4}
\providecommand{\BIBentryALTinterwordspacing}{\spaceskip=\fontdimen2\font plus
\BIBentryALTinterwordstretchfactor\fontdimen3\font minus
  \fontdimen4\font\relax}
\providecommand{\BIBforeignlanguage}[2]{{%
\expandafter\ifx\csname l@#1\endcsname\relax
\typeout{** WARNING: IEEEtran.bst: No hyphenation pattern has been}%
\typeout{** loaded for the language `#1'. Using the pattern for}%
\typeout{** the default language instead.}%
\else
\language=\csname l@#1\endcsname
\fi
#2}}
\providecommand{\BIBdecl}{\relax}
\BIBdecl

\bibitem{azadi2018multi}
S.~Azadi, M.~Fisher, V.~G. Kim, Z.~Wang, E.~Shechtman, and T.~Darrell,
  ``Multi-content gan for few-shot font style transfer,'' in \emph{IEEE
  Conference on Computer Vision and Pattern Recognition}, 2018, pp. 7564--7573.

\bibitem{wu2019editing}
L.~Wu, C.~Zhang, J.~Liu, J.~Han, J.~Liu, E.~Ding, and X.~Bai, ``Editing text in
  the wild,'' in \emph{ACM International Conference on Multimedia}, 2019, pp.
  1500--1508.

\bibitem{kao2019patch}
C.-C. Kao, Y.~Wang, J.~Waltman, and P.~Sen, ``Patch-based image hallucination
  for super resolution with detail reconstruction from similar sample images,''
  \emph{IEEE Transactions on Multimedia}, vol.~22, no.~5, pp. 1139--1152, 2019.

\bibitem{liu2020erasenet}
C.~Liu, Y.~Liu, L.~Jin, S.~Zhang, C.~Luo, and Y.~Wang, ``Erasenet: End-to-end
  text removal in the wild,'' \emph{IEEE Transactions on Image Processing},
  vol.~29, pp. 8760--8775, 2020.

\bibitem{tursun2019mtrnet}
O.~Tursun, R.~Zeng, S.~Denman, S.~Sivapalan, S.~Sridharan, and C.~Fookes,
  ``Mtrnet: A generic scene text eraser,'' in \emph{IAPR International
  Conference on Document Analysis and Recognition}, 2019, pp. 39--44.

\bibitem{tursun2020mtrnet++}
O.~Tursun, S.~Denman, R.~Zeng, S.~Sivapalan, S.~Sridharan, and C.~Fookes,
  ``Mtrnet++: One-stage mask-based scene text eraser,'' \emph{Computer Vision
  and Image Understanding}, vol. 201, p. 103066, 2020.

\bibitem{tang2021stroke}
Z.~Tang, T.~Miyazaki, Y.~Sugaya, and S.~Omachi, ``Stroke-based scene text
  erasing using synthetic data for training,'' \emph{IEEE Transactions on Image
  Processing}, vol.~30, pp. 9306--9320, 2021.

\bibitem{gupta2016synthetic}
A.~Gupta, A.~Vedaldi, and A.~Zisserman, ``Synthetic data for text localisation
  in natural images,'' in \emph{IEEE Conference on Computer Vision and Pattern
  Recognition}, 2016, pp. 2315--2324.

\bibitem{goodfellow2014generative}
I.~Goodfellow, J.~Pouget-Abadie, M.~Mirza, B.~Xu, D.~Warde-Farley, S.~Ozair,
  A.~Courville, and Y.~Bengio, ``Generative adversarial nets,'' \emph{Neural
  Information Processing Systems}, vol.~27, 2014.

\bibitem{nakamura2017scene}
T.~Nakamura, A.~Zhu, K.~Yanai, and S.~Uchida, ``Scene text eraser,'' in
  \emph{IAPR International Conference on Document Analysis and Recognition},
  2017, pp. 832--837.

\bibitem{zhang2019ensnet}
S.~Zhang, Y.~Liu, L.~Jin, Y.~Huang, and S.~Lai, ``Ensnet: Ensconce text in the
  wild,'' in \emph{AAAI Conference on Artificial Intelligence}, 2019, pp.
  801--808.

\bibitem{mirza2014conditional}
M.~Mirza and S.~Osindero, ``Conditional generative adversarial nets,''
  \emph{arXiv:1411.1784}, 2014.

\bibitem{zdenek2020erasing}
J.~Zdenek and H.~Nakayama, ``Erasing scene text with weak supervision,'' in
  \emph{IEEE Winter Conference on Applications of Computer Vision}, 2020, pp.
  2238--2246.

\bibitem{korbar2018cooperative}
B.~Korbar, D.~Tran, and L.~Torresani, ``Cooperative learning of audio and video
  models from self-supervised synchronization,'' \emph{Neural Information
  Processing Systems}, vol.~31, 2018.

\bibitem{fernando2017self}
B.~Fernando, H.~Bilen, E.~Gavves, and S.~Gould, ``Self-supervised video
  representation learning with odd-one-out networks,'' in \emph{IEEE Conference
  on Computer Vision and Pattern Recognition}, 2017, pp. 3636--3645.

\bibitem{buchler2018improving}
U.~Buchler, B.~Brattoli, and B.~Ommer, ``Improving spatiotemporal
  self-supervision by deep reinforcement learning,'' in \emph{European
  Conference on Computer Vision}, 2018, pp. 770--786.

\bibitem{huang2020learning}
Y.~Huang, S.~Qiu, C.~Wang, and C.~Li, ``Learning representations for
  high-dynamic-range image color transfer in a self-supervised way,''
  \emph{IEEE Transactions on Multimedia}, vol.~23, pp. 176--188, 2020.

\bibitem{doersch2015unsupervised}
C.~Doersch, A.~Gupta, and A.~A. Efros, ``Unsupervised visual representation
  learning by context prediction,'' in \emph{IEEE Conference on Computer Vision
  and Pattern Recognition}, 2015, pp. 1422--1430.

\bibitem{wang2015unsupervised}
X.~Wang and A.~Gupta, ``Unsupervised learning of visual representations using
  videos,'' in \emph{International Conference on Computer Vision}, 2015, pp.
  2794--2802.

\bibitem{zhang2016colorful}
R.~Zhang, P.~Isola, and A.~A. Efros, ``Colorful image colorization,'' in
  \emph{European Conference on Computer Vision}, 2016, pp. 649--666.

\bibitem{pathak2016context}
D.~Pathak, P.~Krahenbuhl, J.~Donahue, T.~Darrell, and A.~A. Efros, ``Context
  encoders: Feature learning by inpainting,'' in \emph{IEEE Conference on
  Computer Vision and Pattern Recognition}, 2016, pp. 2536--2544.

\bibitem{gidaris2018unsupervised}
S.~Gidaris, P.~Singh, and N.~Komodakis, ``Unsupervised representation learning
  by predicting image rotations,'' in \emph{International Conference on
  Learning Representations}, 2018.

\bibitem{dosovitskiy2014discriminative}
A.~Dosovitskiy, J.~T. Springenberg, M.~Riedmiller, and T.~Brox,
  ``Discriminative unsupervised feature learning with convolutional neural
  networks,'' \emph{Neural Information Processing Systems}, vol.~27, 2014.

\bibitem{wang2019shape}
W.~Wang, E.~Xie, X.~Li, W.~Hou, T.~Lu, G.~Yu, and S.~Shao, ``Shape robust text
  detection with progressive scale expansion network,'' in \emph{IEEE
  Conference on Computer Vision and Pattern Recognition}, 2019, pp. 9336--9345.

\bibitem{zhan2019esir}
F.~Zhan and S.~Lu, ``Esir: end-to-end scene text recognition via iterative
  rectification,'' in \emph{IEEE Conference on Computer Vision and Pattern
  Recognition}, 2019, pp. 2059--2068.

\bibitem{gao2020progressive}
Y.~Gao, Y.~Chen, J.~Wang, and H.~Lu, ``Progressive rectification network for
  irregular text recognition,'' \emph{Science China Information Sciences},
  vol.~63, no.~2, pp. 1--14, 2020.

\bibitem{chen2020progressively}
S.~Chen and Y.~Fu, ``Progressively guided alternate refinement network for
  rgb-d salient object detection,'' in \emph{European Conference on Computer
  Vision}, 2020, pp. 520--538.

\bibitem{saharia2021image}
C.~Saharia, J.~Ho, W.~Chan, T.~Salimans, D.~J. Fleet, and M.~Norouzi, ``Image
  super-resolution via iterative refinement,'' \emph{arXiv:2104.07636}, 2021.

\bibitem{wang2021pert}
Y.~Wang, H.~Xie, S.~Fang, Y.~Qu, and Y.~Zhang, ``Pert: A progressively
  region-based network for scene text removal,'' \emph{arXiv:2106.13029}, 2021.

\bibitem{he2016deep}
K.~He, X.~Zhang, S.~Ren, and J.~Sun, ``Deep residual learning for image
  recognition,'' in \emph{IEEE Conference on Computer Vision and Pattern
  Recognition}, 2016, pp. 770--778.

\bibitem{isola2017image}
P.~Isola, J.-Y. Zhu, T.~Zhou, and A.~A. Efros, ``Image-to-image translation
  with conditional adversarial networks,'' in \emph{IEEE Conference on Computer
  Vision and Pattern Recognition}, 2017, pp. 1125--1134.

\bibitem{gatys2015neural}
L.~A. Gatys, A.~S. Ecker, and M.~Bethge, ``A neural algorithm of artistic
  style,'' \emph{arXiv:1508.06576}, 2015.

\bibitem{iizuka2017globally}
S.~Iizuka, E.~Simo-Serra, and H.~Ishikawa, ``Globally and locally consistent
  image completion,'' \emph{ACM Transactions on Graphics}, vol.~36, no.~4, pp.
  1--14, 2017.

\bibitem{yu2019free}
J.~Yu, Z.~Lin, J.~Yang, X.~Shen, X.~Lu, and T.~S. Huang, ``Free-form image
  inpainting with gated convolution,'' in \emph{International Conference on
  Computer Vision}, 2019, pp. 4471--4480.

\bibitem{miyato2018spectral}
T.~Miyato, T.~Kataoka, M.~Koyama, and Y.~Yoshida, ``Spectral normalization for
  generative adversarial networks,'' in \emph{International Conference on
  Learning Representations}, 2018.

\bibitem{baek2019character}
Y.~Baek, B.~Lee, D.~Han, S.~Yun, and H.~Lee, ``Character region awareness for
  text detection,'' in \emph{IEEE Conference on Computer Vision and Pattern
  Recognition}, 2019, pp. 9365--9374.

\bibitem{liu2019curved}
Y.~Liu, L.~Jin, S.~Zhang, C.~Luo, and S.~Zhang, ``Curved scene text detection
  via transverse and longitudinal sequence connection,'' \emph{Pattern
  Recognition}, vol.~90, pp. 337--345, 2019.

\bibitem{shi2017icdar2017}
B.~Shi, C.~Yao, M.~Liao, M.~Yang, P.~Xu, L.~Cui, S.~Belongie, S.~Lu, and
  X.~Bai, ``Icdar2017 competition on reading chinese text in the wild
  (rctw-17),'' in \emph{IAPR International Conference on Document Analysis and
  Recognition}, vol.~1, 2017, pp. 1429--1434.

\bibitem{ch2017total}
C.~K. Ch'ng and C.~S. Chan, ``Total-text: A comprehensive dataset for scene
  text detection and recognition,'' in \emph{IAPR International Conference on
  Document Analysis and Recognition}, vol.~1, 2017, pp. 935--942.

\bibitem{yao2012detecting}
C.~Yao, X.~Bai, W.~Liu, Y.~Ma, and Z.~Tu, ``Detecting texts of arbitrary
  orientations in natural images,'' in \emph{IEEE Conference on Computer Vision
  and Pattern Recognition}, 2012, pp. 1083--1090.

\bibitem{paszke2019pytorch}
A.~Paszke, S.~Gross, F.~Massa, A.~Lerer, J.~Bradbury, G.~Chanan, T.~Killeen,
  Z.~Lin, N.~Gimelshein, L.~Antiga, A.~Desmaison, A.~Kopf, E.~Yang, Z.~DeVito,
  M.~Raison, A.~Tejani, S.~Chilamkurthy, B.~Steiner, L.~Fang, J.~Bai, and
  S.~Chintala, ``Pytorch: An imperative style, high-performance deep learning
  library,'' in \emph{Neural Information Processing Systems}, vol.~32, 2019.

\bibitem{kingma2014adam}
D.~P. Kingma and J.~Ba, ``Adam: A method for stochastic optimization,'' in
  \emph{International Conference on Learning Representations}, 2015.

\end{thebibliography}

\end{document}